%% file: main.tex
\documentclass[runningheads]{llncs}
\usepackage{graphicx}
\usepackage{comment}
\usepackage{amsmath,amssymb} %
\usepackage{color}
\usepackage{physics}

\usepackage{xspace}
\usepackage{multirow}
\usepackage{bbm}

\usepackage{multirow}
\usepackage{nopageno}

\usepackage{algorithm}
\usepackage[noend]{algpseudocode}
\usepackage{tabularx}
\usepackage{makecell}
\usepackage{microtype}
\usepackage{booktabs}
\usepackage{textcomp}
\usepackage{colortbl}
\usepackage[table]{xcolor}
\usepackage{romannum}
\usepackage{url}
\usepackage{wrapfig}

\usepackage[width=122mm,left=12mm,paperwidth=146mm,height=193mm,top=12mm,paperheight=217mm]{geometry}

\input{define.tex}

\input{kaidefine}
\begin{document}
\pagestyle{headings}
\mainmatter
\def\ECCVSubNumber{736}  %

\title{Crowdsampling the Plenoptic Function} %

\titlerunning{Crowdsampling the Plenoptic Function}
\author{Zhengqi Li \and
Wenqi Xian\and
Abe Davis \and
Noah Snavely}

\authorrunning{Li et al.}
\institute{Cornell Tech, Cornell University}

\maketitle

\input{00-abstract}

\input{01-intro}

\input{02-related}
\input{03-method}
\input{04-experiments}

\input{06-conclusion}

\bibliographystyle{splncs04}
\bibliography{refs}
\end{document}

%% file: define.tex
\input{abedefine}

\newcommand{\lfu}{u}
\newcommand{\lfv}{v}
\newcommand{\lfx}{x}
\newcommand{\lfy}{y}

\newcommand{\im}[1]{I_{#1}}

\newcommand{\tRv}{$r$}

\newcommand{\LF}{\mathbf{\mathcal{L}}}

\newcommand{\LFf}[1]{\LF{}(#1)}
\newcommand{\tLFf}[1]{$\LFf{#1}$}

\newcommand{\PF}[1]{\mathbf{\mathcal{P}}}
\newcommand{\tPF}[1]{$\PF$}

\newcommand{\wgan}{w_{\mathsf{GAN}}}
\newcommand{\wstyle}{w_{\mathsf{style}}}

\newcommand{\Lvgg}{\mathcal{L}_{\mathsf{VGG}}}
\newcommand{\Lgan}{\mathcal{L}_{\mathsf{GAN}}}
\newcommand{\Lstyle}{\mathcal{L}_{\mathsf{style}}}

\newcommand{\rfmpi}{D^r}

\newcommand{\kfmpi}{D^k}

\newcommand{\predmpi}{{{C}}^k_s}

\newcommand{\etal}{\textit{et al.}}

\newcommand{\deepmpi}{DeepMPI\xspace}

\newcommand{\albedo}{base color\xspace}
\newcommand{\albedok}{\hat{B}^k}
\newcommand{\albedor}{B^r}
\newcommand{\deepbuffer}{\Phi_s^r}
\newcommand{\lv}{\mathbf{z}_s}

\newcommand{\sourcei}{I_s}

%% file: abedefine.tex
\usepackage[strict]{changepage}

\usepackage{framed}

\definecolor{formalshade}{rgb}{0.95,0.95,0.95}

\newenvironment{formalblock}{%
  \MakeFramed{\advance\hsize-\width\FrameRestore}%
  \noindent\hspace{-4.55pt}%
  \begin{adjustwidth}{}{7pt}%
  \vspace{2pt}\vspace{2pt}%
  \footnotesize
}
{%
  \vspace{2pt}\end{adjustwidth}\endMakeFramed%
}

%% file: 00-abstract.tex
\begin{abstract}
Many popular tourist landmarks are captured in a multitude of online, public photos. These photos represent a sparse and unstructured sampling of the plenoptic function for a particular scene. In this paper, we present a new approach to novel view synthesis under time-varying illumination from such data. 
Our approach builds on the recent \emph{multi-plane image} (MPI) format for representing local light fields under fixed viewing conditions. We introduce a new \textit{\deepmpi} representation, motivated by observations on the sparsity structure of the plenoptic function, that allows for real-time synthesis of photorealistic views that are continuous in both space and across changes in lighting. Our method can synthesize the same compelling parallax and view-dependent effects as previous MPI methods, while simultaneously interpolating along changes in reflectance and illumination with time. We show how to learn a model of these effects in an unsupervised way from an unstructured collection of photos without temporal registration, demonstrating significant improvements over recent work in neural rendering. More information can be found at \url{crowdsampling.io}.
\end{abstract}

%% file: 01-intro.tex
\section{Introduction}

\begin{figure}[t]
\centering
  \includegraphics[width=\columnwidth]{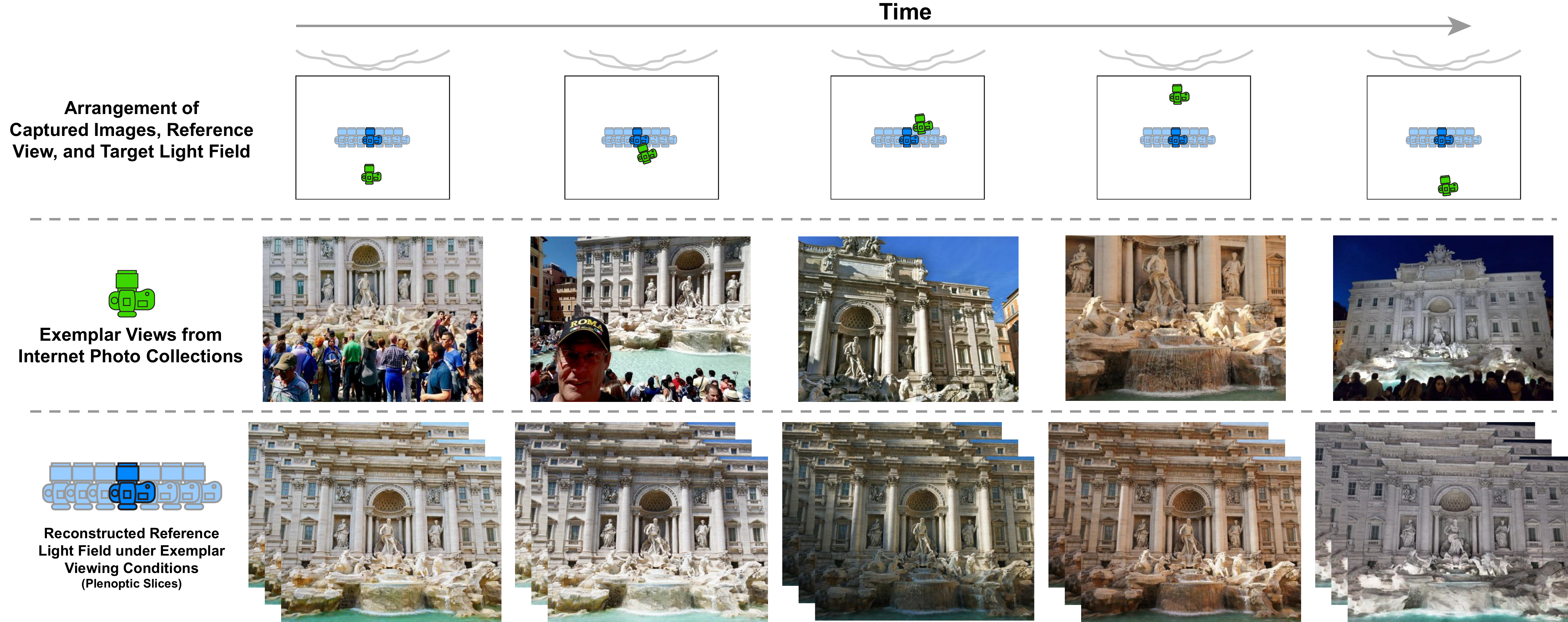}
  \caption{\textbf{Crowdsampled plenoptic slices.}  Given a large number of tourist photos taken at different times of day, our system learns to construct a continuous set of lightfields and to synthesize novel views capturing all-times-of-day scene appearance.}
\label{fig:teaser}
\end{figure}

There is a thought experiment that goes something like this:
\begin{formalblock}
Imagine a `camera' with no optics or image sensor of any kind. Rather, it consists only of a box equipped with GPS, a radio for Internet access, a button for `taking pictures', and a screen for displaying those pictures. When a user presses its button, the box searches the Internet for photos tagged with its current location, and from these selects a best match to display on the screen.
\end{formalblock}
\noindent This thought experiment is perhaps best understood in the context of popular tourist attractions, of which one can often find countless images posted online (Figure~\ref{fig:teaser}, second row). When pointed at such an attraction, one can imagine our box producing images very similar to those of a real camera, forcing us to consider whether an image we capture ourselves is meaningfully different from a near-identical one captured by strangers.
For many, the ensuing philosophical debate hinges on whether an image reflects the scene as they remember it. After all, appearance is not generally constant over time, even under a fixed geometry and viewpoint; in outdoor settings, for example, weather changes, shadows move, and day turns to night---all resulting in appearance changes that can be observed from a single view of the scene. 

This poses an interesting challenge to the field of image-based rendering: can we use crowdsourced imagery to synthesize arbitrary views of a scene with viewing conditions that change over time?
Without changing viewing conditions, this challenge would reduce to the more familiar problem of reconstructing a 4D light field \tLFf{\lfu,\lfv,\lfx,\lfy} that describes all light in our scene \cite{levoy1996light}.
When we add time to our problem, it turns into a 5D reconstruction over what Adelson and Bergen \cite{Adelson91theplenoptic} call the \emph{plenoptic function}.\footnote{
\cite{Adelson91theplenoptic} describes the plenoptic function as 7D, but we can reduce this to a 4D color light field supplemented by time by applying the later observations of \cite{levoy1996light}.
}

In this paper, we propose a novel approach to neural image-based rendering from 
crowdsourced images that leverages 
the sparse structure of the plenoptic function to learn how scene appearance changes over space and time in an unsupervised manner. 
Our approach takes unstructured Internet photos spanning some range of time-varying appearance in a scene and learns how to reconstruct a \emph{plenoptic slice}---a representation of the light field that respects temporal structure in the plenoptic function when interpolated over time---for each of the viewing conditions captured in our input data. 
By designing our model to preserve the structure of real plenoptic functions, we force it to learn time-varying phenomena like the motion of shadows according to sun position. This lets us, for example, recover plenoptic slices for images taken at different times of day (Figure~\ref{fig:teaser}, bottom row) and interpolate between them to observe how shadows move as the day progresses (best seen in our supplemental video). In effect, we learn a representation of the scene that can produce high-quality views from a continuum of viewpoints \textit{and} viewing conditions that vary with time.

Our work makes three key contributions: first, 
a representation, called a DeepMPI, for neural rendering that extends prior work on multiplane images (MPIs) \cite{zhou2018stereo} to model viewing conditions that vary with time; second, 
a method for training DeepMPIs on sparse,  unstructured crowdsampled data that is unregistered in time; and third, 
a dataset of crowdsampled images taken from Internet photo collections, along with details on how it was collected and registered.

Compared with previous work, our approach inherits the advantages of recent methods based on MPIs~\cite{zhou2018stereo,srinivasan2019pushing,mildenhall2019local,choi2019extreme,flynn2019deepview}, including the ability to produce high-quality novel views of complex scenes in real time and the view consistency that arises from a 3D scene representation (in contrast to neural rendering approaches that decode a separate view for each desired viewpoint). To these advantages we add the key ability to synthesize and interpolate continuous, photo-realistic, time-varying changes in appearance. We compare our approach both quantitatively and qualitatively to recent neural rendering methods, such as Neural Rerendering in the Wild~\cite{Meshry2019NeuralRI}, and show that our method produces superior results.

%% file: 02-related.tex
\section{Related Work}

The study of image-based rendering is motivated by a simple question: how do we use a finite set of images to reconstruct an infinite set of views? Different branches of research have explored this question from different angles and with different assumptions.
Here we outline the space of approaches,
highlighting work most closely related to our own.

\medskip
\noindent\textbf{Novel view synthesis.}
Novel view synthesis has traditionally been approached through either explicit estimation of scene geometry and color~\cite{Hedman2017Casual3P,Zitnick2004HighqualityVV,chaurasia2013depth}, or using coarser estimates of geometry to guide interpolation between captured views~\cite{buehler2001unstructured,debevec1996modeling,snavely2006phototourism}. Light field rendering~\cite{levoy1996light,gortler1996lumigraph,plenopticsampling00} pushes the latter strategy to an extreme by using dense structured sampling of the light field to make reconstruction guarantees independent of specific scene geometry. Subsequent works~\cite{Levin2010LinearVS,Shi2014LightFR,Davis2012UnstructuredLF,Vagharshakyan2015LightFR,penner2017soft,sparselightfields14} have leveraged observations on the structure of light fields to build on this approach.
However, most IBR algorithms are designed to model static appearance, making them ill-suited for our problem.

Recently, deep learning techniques have been applied to this problem. 
Several works~\cite{thies2019deferred,hedman2018deep} rely on global meshes to guide view synthesis. However, such methods heavily rely on the accuracy of 3D models, and often fail to model complex scene components such as translucent and thin objects. Other works 
predict appearance flow~\cite{zhou2016view}, depth probabilities~\cite{flynn2016deepstereo,xu2019deepviewsyn}, or RGBD light fields~\cite{kalantari2016learning,srinivasan2017learning}. However, many of these methods independently synthesize appearance for each view, leading to inconsistent renderings across views.

Our approach builds on the use of multiplane images (MPIs)~\cite{zhou2018stereo} for novel view synthesis.
Several recent methods have shown that MPIs are an effective and learnable representation for light fields~\cite{srinivasan2019pushing,mildenhall2019local,choi2019extreme,flynn2019deepview}. We build on this representation by introducing the \deepmpi, which further captures 
viewing condition--dependent appearance. We are also inspired by recent work that poses view synthesis as decoding features from a learned latent space~\cite{sitzmann2019deepvoxels,lombardi2019neural,sitzmann2019scene,thies2019deferred,chen2019neural,eslami2018neural}. However, such work has been limited to synthetic environments or objects captured in controlled settings and is difficult to apply to crowdsampled images.

\medskip
\noindent\textbf{Appearance modeling.} 
Several works have modeled the time-varying appearance of outdoor scenes using physically-motivated approaches~\cite{Shan2013TheVT,Hauagge2014ReasoningAP,Laffont2012CoherentII} or by combining data-driven methods and dense geometry~\cite{garg2009dimensionality,philip2019multi,yu2019inverserendernet,matzen2014scene}. Additionally, Martin-Brualla~\etal~\cite{martin2015time,martin20153d} reconstruct time-lapses of urban scenes from Internet photos. However, their method relies on timestamps, and models appearance changes at much coarser granularity (scene dynamics across years). The recent work of Meshry~\etal~\cite{Meshry2019NeuralRI} is probably closest to our own. They model appearance changes across varying times of day by learning an appearance embedding. However, their method relies heavily on dense multi-view stereo geometry,
and tends to produce temporal artifacts under complex appearance changes. In contrast, our approach is capable of rendering a more continuous range of photo-realistic views across diverse appearances, without relying on dense input geometry.

\medskip
\noindent\textbf{Deep image synthesis.}
Our work is also related to the problem of image-to-image translation~\cite{chen2017photographic,isola2017image,wang2018high,park2019semantic}, multi-model image-to-image translation~\cite{zhu2017unpaired,zhu2017toward,huang2018multimodal,lee2018diverse} and style transfer~\cite{gatys2016image,ulyanov2017improved,huang2017arbitrary,sheng2018avatar}. Recently, Generative Adversarial Networks (GANs)~\cite{goodfellow2014generative,mao2017least,gulrajani2017improved} have successfully produced photo-realistic imagery, enabling a variety of applications in deep image synthesis~\cite{xian2018texturegan,sangkloy2017scribbler,karras2019style,karras2017progressive,ledig2017photo,wang2016generative}. However, there has been comparatively little investigation of 3D scene representations for deep image synthesis. Our method demonstrates the ability to learn a generative 3D scene representation and produce high-quality novel views of complex scenes. 

%% file: 03-method.tex
\section{Approach}

Given a set $\mathcal{I} =\{\im{1},\im{2},...,\im{n}\}$ of crowdsampled photos with corresponding camera viewpoints $\mathcal{C} = \{c_1, c_2,..., c_n\}$ captured in a common scene, we formulate our problem as the reconstruction of \textit{plenoptic slices} (local light fields parameterized by an apperance descriptor) around some reference view $r$ conditioned on each of the scene appearances captured in $\mathcal{I}$ (see Figure \ref{fig:teaser} for a geometric sketch of this setup). 
We present our approach in three parts: first, we describe how the input images $\mathcal{I}$ are collected and registered (Section \ref{sec:dataset}); then we discuss our representation of the plenoptic function, which extends multiplane images (MPIs) to model 
appearance changes over time (Section \ref{sec:representation}); and finally we describe how to train this representation on our crowdsampled data (Sections \ref{sec:mpi} and \ref{sec:viewsysn}).

\smallskip
\noindent \textit{Note on notation:} Throughout the paper, we will use superscripts to denote camera viewpoints and subscripts to denote image or voxel indices.

\subsection{Collecting Crowdsampled Data} \label{sec:dataset}

We 
selected a number of popular tourist sites and downloaded $\sim$50K photos from Flickr for each site. For each scene, we must then register these photos by solving for a camera pose and intrinsic parameters for each image. As running structure from motion (SfM) from scratch on such quantities of images is very expensive, we instead started with a existing SfM reconstruction of each site from the MegaDepth dataset~\cite{li2018megadepth}, and performed camera relocalization 
to efficiently register each new image against the existing reconstruction~\cite{schonberger2016structure}. 

For each landmark, we then identified a reference viewpoint \tRv{} to center our reconstruction by using a canonical view selection algorithm similar to that of Simon \etal
to find viewpoints with a high density of nearby views~\cite{simon2007scene}. We then select all images captured from within a sphere centered at \tRv{} for use in our method, randomly splitting the set gathered from each landmark into training and test data.
We manually set the field of view of the reference viewpoint so that it has good coverage of the scene.

\begin{figure}[t!]
  \centering
    \begin{tabular}{@{}c@{}c@{}c@{}}
        \includegraphics[width=0.35\columnwidth]{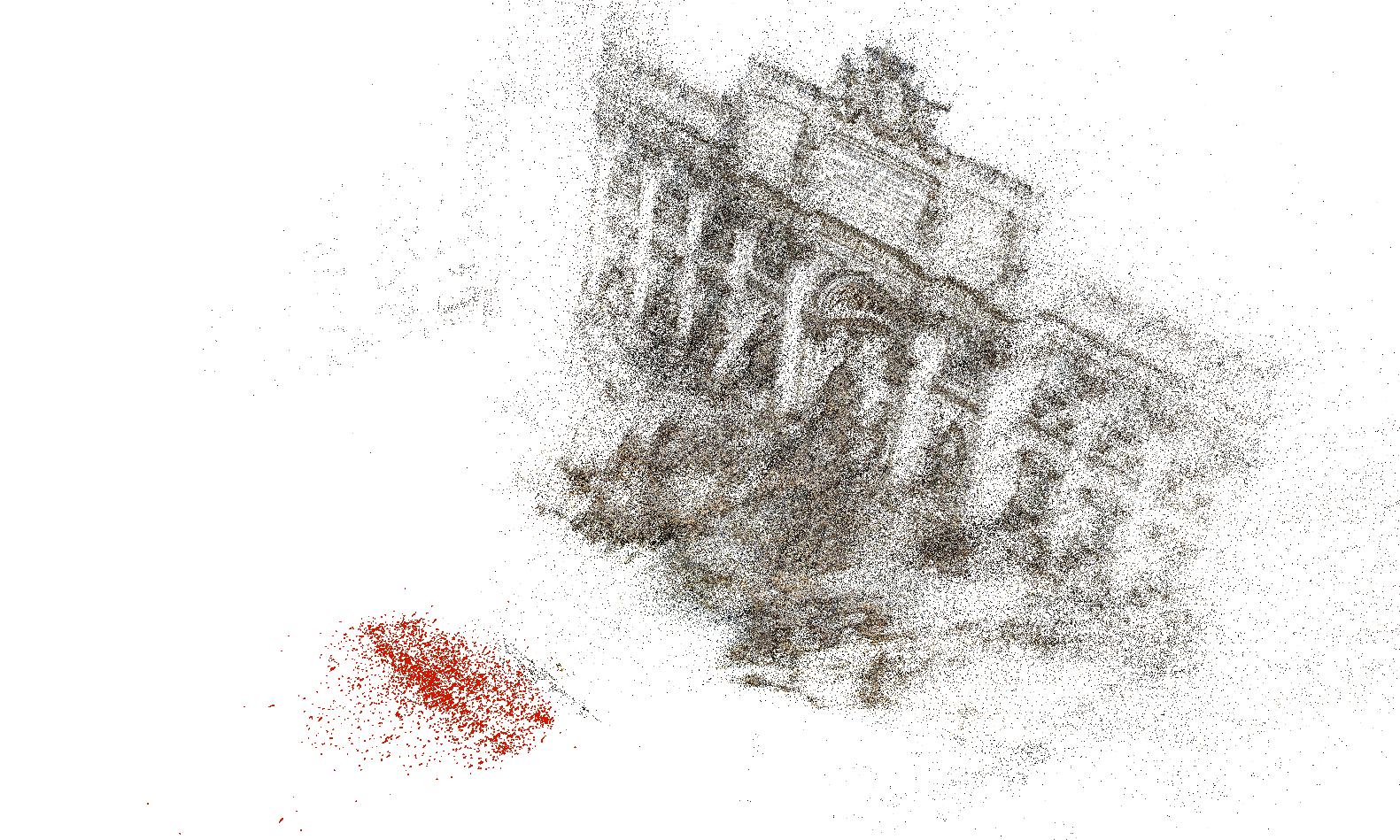} & 
        \includegraphics[width=0.35\columnwidth]{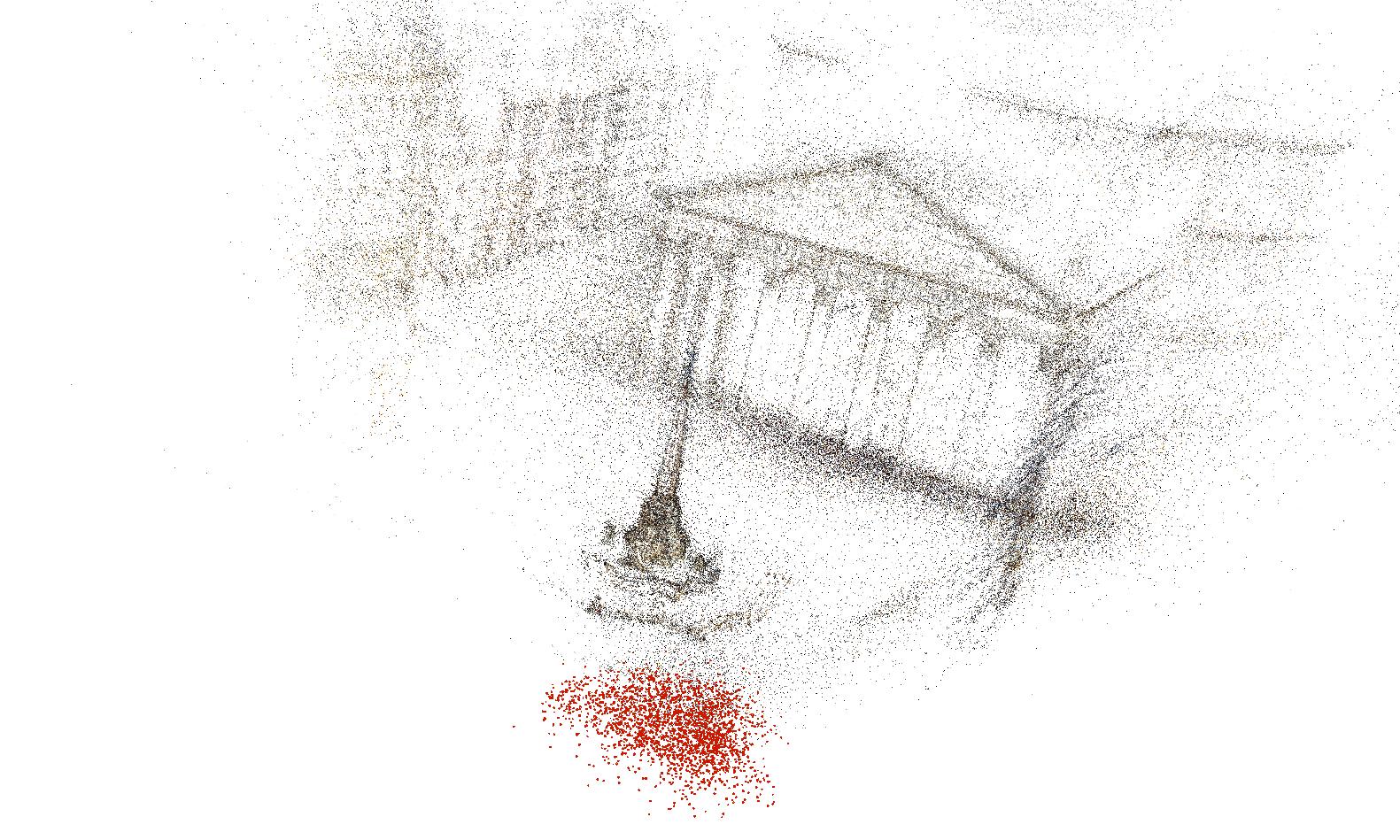} &
        \includegraphics[width=0.35\columnwidth]{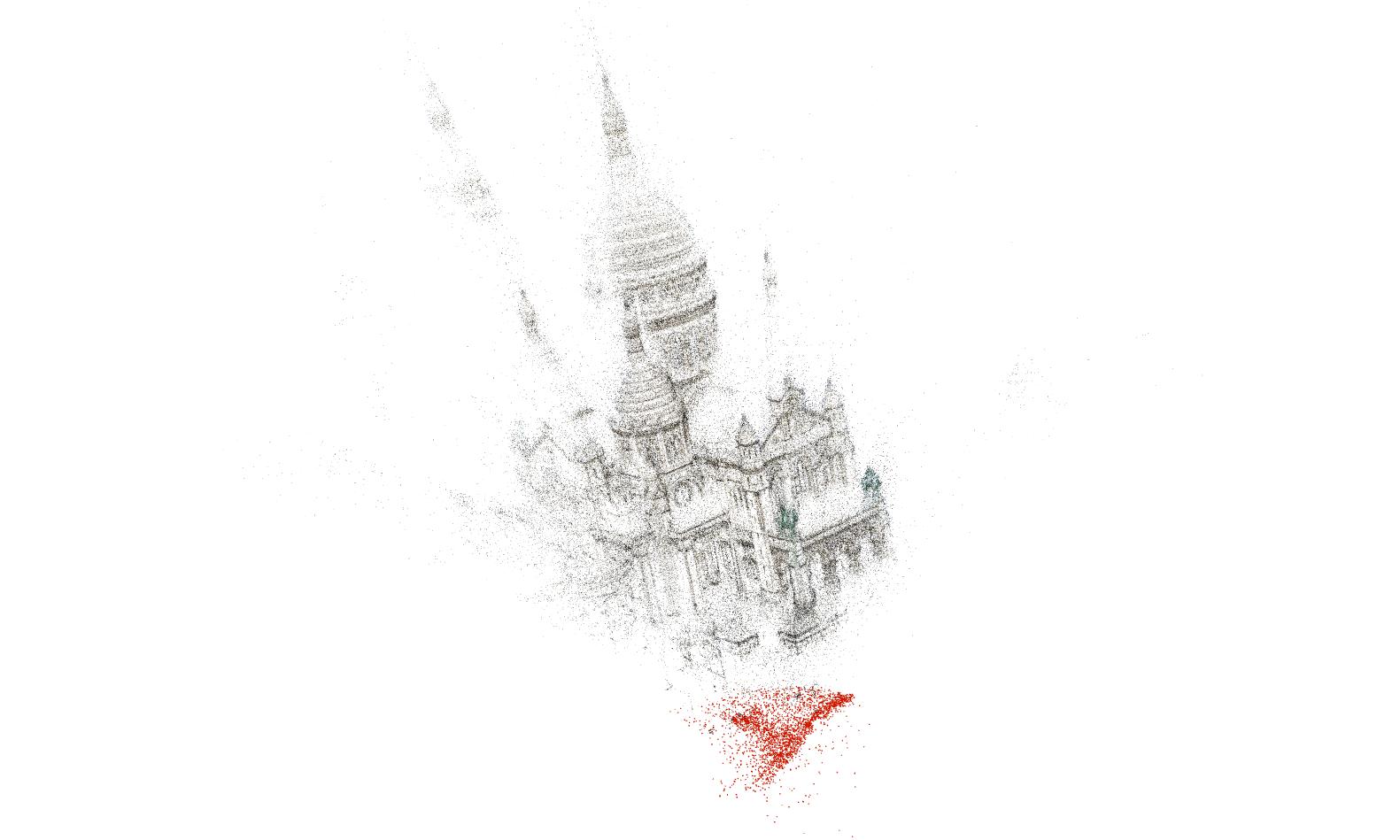} \\
        (a) Trevi Fountain  & 
        (b) The Pantheon &
        (c) Sacre Coeur \\
    \end{tabular}
  	\caption{\textbf{Registered photo collections}. Example SfM reconstructions of clusters of Internet photos sharing similar viewpoints, labeled as red dots. \label{fig:dataset}}
 \end{figure}

We found that the camera parameters estimated from relocalization are sometimes inaccurate, and so we reapply a global SfM and bundle adjustment to the smaller set of selected images near each scene's reference view to reestimate these images' camera parameters. We used this data pipeline to gather and register photos for eight locations, and will release this data to the research community. Figure~\ref{fig:dataset} shows final SfM reconstructions for three of these landmarks.

\subsection{The \deepmpi Scene Representation} \label{sec:representation}

We base our representation on the multiplane image (MPI) format~\cite{szeliski98stereomatching,zhou2018stereo}, which represents light fields locally as a stack of fronto-parallel planar RGB$\alpha$ layers arranged at varying distances from the camera, akin to a stack of transparencies. Novel views are rendered from an MPI by warping the layers into a new view, then performing an \emph{over} operation to composite the warped layers into a rendered image. Individual RGB$\alpha$ elements (``voxels'') of an MPI are indexed by $(x,y)$ position and plane depth $d$.

While MPIs have been 
remarkably effective for reconstructing fixed light fields from sparse views~\cite{flynn2019deepview}, they do not encode any information about how viewing conditions may vary with time. Furthermore, even if we were given a regular MPI corresponding to viewing conditions for each of our input images, directly interpolating between these MPIs would still fail to capture temporal structure in the plenoptic function. For example, interpolating between morning and afternoon MPIs would cause shadows cast by the sun to appear in duplicate when, in reality, a single shadow moved over time. This observation highlights the distinction between what we call a light field and what we call a plenoptic slice: we use the latter to describe a reparameterization of the light field that is better-suited for interpolation over time.

Inspired by DeepVoxels~\cite{sitzmann2019deepvoxels}, we introduce \emph{DeepMPIs} to help learn this reparameterization.
DeepMPIs augment standard RGB$\alpha$ MPIs by appending a \emph{learnable} latent feature vector at each MPI voxel (see Figure~\ref{fig:pipeline}). For a given scene, we position a \deepmpi at the reference viewpoint $r$, and denote this reference \deepmpi as $\rfmpi = (\albedor, \alpha^r, F^r)$. Each voxel of $\rfmpi$ at spatial location and depth $\mathbf{p} = (x, y, d)$
consists of a base RGB color $B^r_\mathbf{p}$, an alpha weight $\alpha^r_\mathbf{p}$, and a latent feature vector $F^r_\mathbf{p}$. 
We set the number of \deepmpi depth planes to $64$ with uniform sampling in disparity space, and we adopt the method of Zhou~\etal~\cite{zhou2018stereo} to set the depth of the near and far planes of the \deepmpi.

In our supplemental document we relate the design of this representation and its training to priors on the sparse structure of the plenoptic function. At a high level, the $\alpha$ planes encode visibility information, which we expect to remain constant even as lighting and other viewing conditions change with time. The latent feature planes $F^r$ are trained to capture correlations between different viewing conditions that arise from, for example, limited variation in material properties and correlation among surface normals within the scene. 
A plenoptic slice then consists of a \deepmpi and some exemplar image $\im{k}$. We can convert this to a standard RGB$\alpha$ MPI representing appearance under the specific conditions captured in $\im{k}$ by using a decoder that is trained jointly with our \deepmpi, which we describe in Section~\ref{sec:viewsysn}.

To compute a \deepmpi from a collection of registered images, we use a two-stage process: first, we first estimate base color and $\alpha$ planes (Section~\ref{sec:mpi}), then optimize latent features $F^r$ jointly with our neural rendering network (Section~\ref{sec:viewsysn}) to enable controllable, varying appearance.

\subsection{Stage 1: Optimizing \deepmpi Color and $\alpha$ Planes} 
\label{sec:mpi}
In the first stage of our method, we optimize \albedo planes $\albedor$ and alpha planes $\alpha^r$ in our \deepmpi as if it were a standard RGB$\alpha$ MPI. One simple approach would be to jointly optimize $\albedor$ and $\alpha^r$ from scratch so as to minimize a reconstruction loss over all images (i.e., the difference between a known image and an MPI-predicted image from that viewpoint, averaged over all input images). However, as described in \cite{flynn2019deepview}, such a method exhibits slow convergence and can be prone to local minima. In addition, compared to~\cite{flynn2019deepview}, our setting is more challenging because Internet photos exhibit diversity in camera parameters and viewing conditions. Instead, we propose a simple yet effective approach to estimating $\albedor$ and $\alpha^r$ given a set of posed input views.

We start by creating a mean RGB plane sweep volume (PSV) at the reference viewpoint by reprojecting every image to the reference viewpoint via each depth plane, then averaging all reprojected images at each depth plane. We initialize the base color planes $\albedor$ to this mean RGB PSV. Keeping these color planes fixed, we optimize the alpha planes $\alpha^r$ to minimize reconstruction losses over the training photos. Specifically, given a photo $I_k$ at viewpoint $c_k$, we project both $B^r$ and $\alpha^r$ to $c_k$, then apply the over operation from back to front to render a base color image $\albedok$: 
\begin{align}
 \albedok = \mathcal{O} \left( \mathcal{W}^k(B^r), \mathcal{W}^k(\alpha^r) \right),    
\end{align}
where $\mathcal{O}$ is the over operation and $\mathcal{W}^k$ is the warping operation from the reference viewpoint \tRv{} to the target viewpoint $c_k$. We compare the rendered \albedo image $\hat{B}^k$ and $I_k$ using a reconstruction loss consisting of a pixel-wise $l_1$ loss and a multi-scale gradient consistency loss~\cite{li2018megadepth,li2019learning}. We  observe that the gradient consistency loss leads to higher rendering quality and faster convergence.%

\begin{figure}[t!]
  \centering
    \begin{tabular}{@{}c@{}c@{}c@{}c@{}}
        \includegraphics[width=0.25\columnwidth]{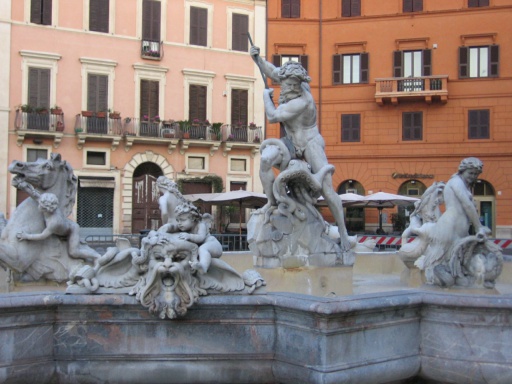} & 
        \includegraphics[width=0.25\columnwidth]{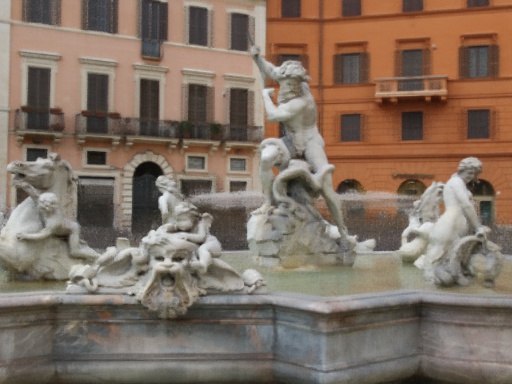} &
        \includegraphics[width=0.25\columnwidth]{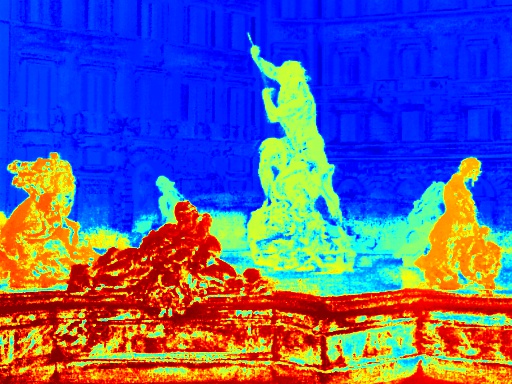} & 
        \includegraphics[width=0.25\columnwidth]{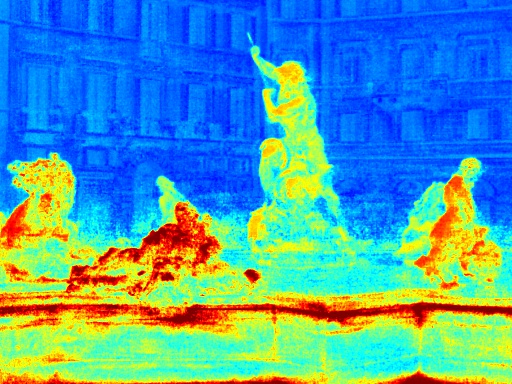} \\
        \includegraphics[width=0.25\columnwidth]{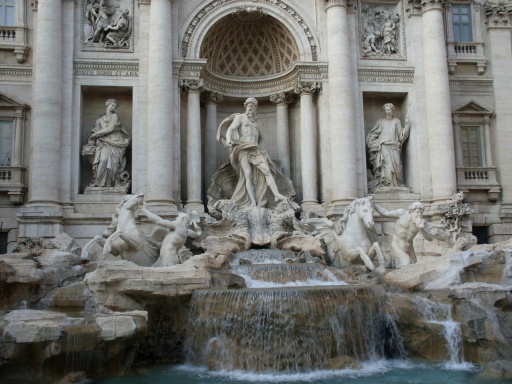} & 
        \includegraphics[width=0.25\columnwidth]{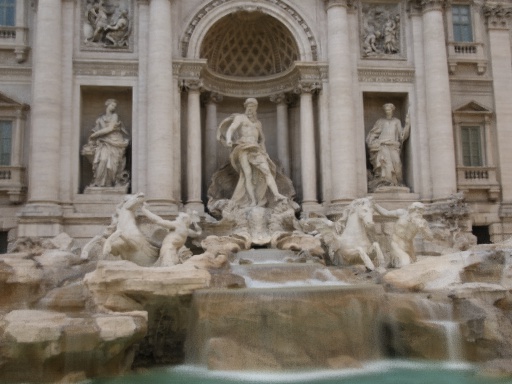} &
        \includegraphics[width=0.25\columnwidth]{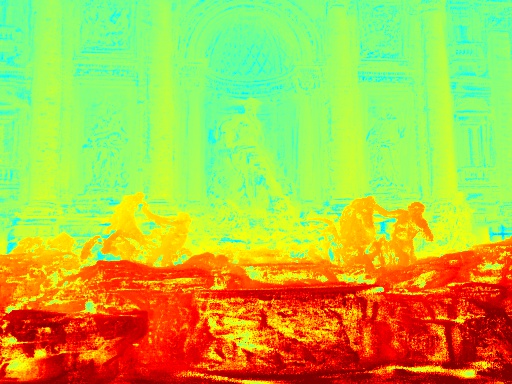} & 
        \includegraphics[width=0.25\columnwidth]{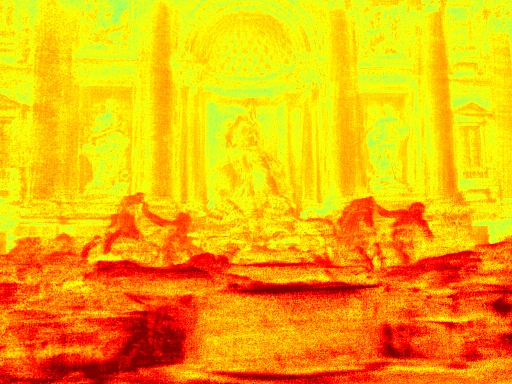} \\
        (a) viewpoint $c_k$ & 
        (b) base color $\hat{B}^k$ &
        (c) our depth &
        (d) baseline depth \\
    \end{tabular}
  	\caption{\textbf{Renderings of \albedo and alpha.} From left to right: (a) original photos at target viewpoint $c_k$, (b) our estimated \albedo at $c_k$, (c) pseudo-depth computed from the RGB$\alpha$ MPI at $c_k$ using our two-phase approach, (d) pseudo-depth from the baseline. For depth maps, red=close and blue=far.  \label{fig:rgba_mpi}}
 \end{figure}

Since the mean RGB PSV cannot accurately model scene content that is occluded in the reference view, after optimizing $\alpha^r$ with fixed $B^r$, we unfreeze $B^r$ and jointly optimize $B^r$ and $\alpha^r$ using the reconstruction loss described above. We observe that this two-phase training method leads to more accurate estimates of $\alpha^r$ than the alternative of optimizing $B^r$ and $\alpha^r$ together from scratch. Figure~\ref{fig:rgba_mpi}, shows examples of input viewpoints and rendered \albedo images, as well as a comparison of pseudo-depths derived from alpha planes $\alpha^r$ computed by our two-phase training method and by the baseline.
Once $\albedor$ and $\alpha^r$ are estimated, they are fixed for the subsequent stage of training, described below.

\subsection{Stage 2: Learning How Appearance Changes with Time} \label{sec:viewsysn}

\begin{figure}[t]
\centering
  \includegraphics[width=\columnwidth]{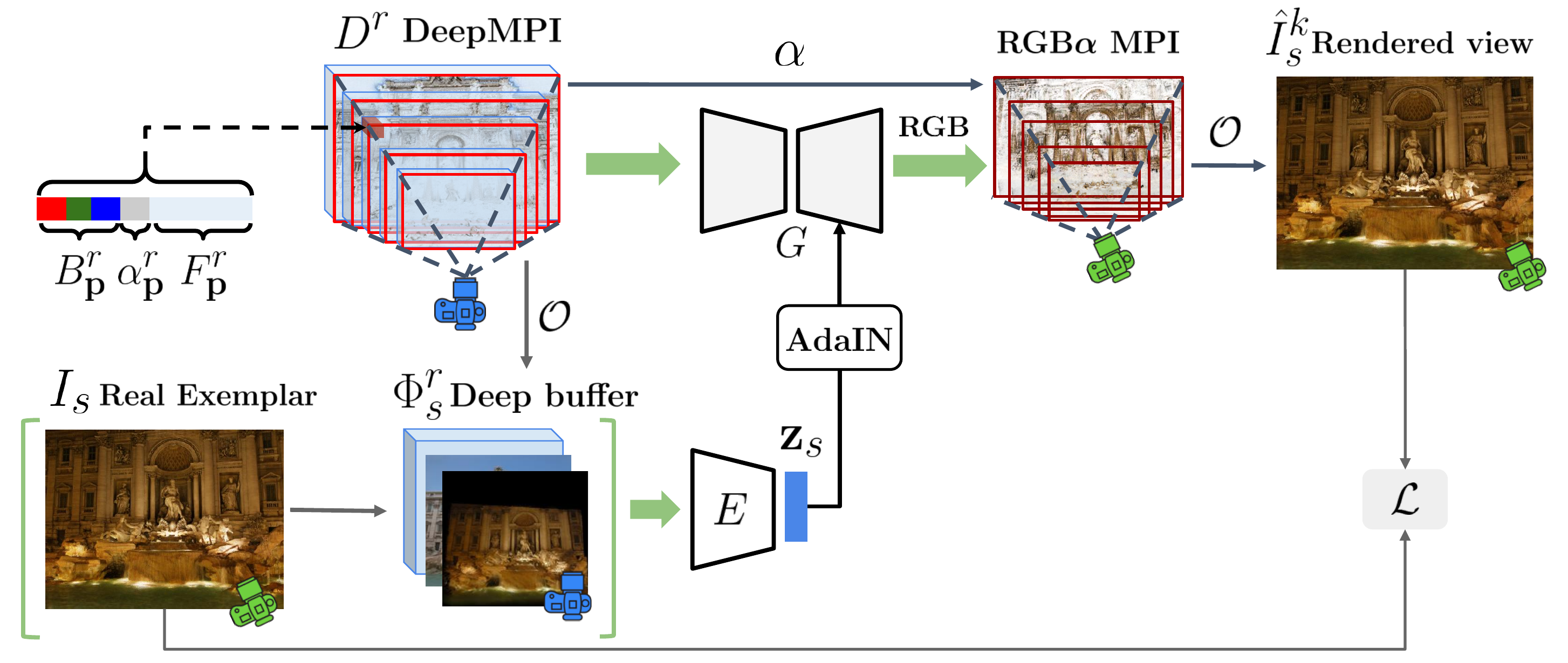}
  \caption{\textbf{Learning framework.} Our method builds a reference \deepmpi $\rfmpi$, consisting of \albedo, alpha, and latent feature components organized into planar layers. A rendering network $G$ takes a \deepmpi projected to a target viewpoint $c_k$, and predicts corresponding RGB color layers. The appearance of these layers is modulated by an appearance vector $\lv$ produced by encoder $E$. The over operation $\mathcal{O}$ is applied to the resulting RGB$\alpha$ MPI to render a view. We jointly train the encoder $E$, rendering network $G$, and latent features $F^r$ in the \deepmpi by comparing a rendered view with an original exemplar image $I_k=I_s$. During inference, given an exemplar photo $I_s$, we can synthesize novel views close to the reference viewpoint, while also preserving the exemplar's appearance.}
    \label{fig:pipeline}
\end{figure}

Our method's second stage optimizes the latent features $F^r$ in our \deepmpi, together with an appearance encoder $E$ and rendering network $G$, to capture and render time-varying appearance. Our learning framework is summarized in Figure~\ref{fig:pipeline}. 

\medskip
\noindent\textbf{Appearance encoder.}
To model appearance variation, we devise a method wherein an encoder $E$ learns to map an exemplar image $\sourcei$ 
and an auxiliary \textit{deep buffer} $\deepbuffer$ 
to a latent appearance vector $\lv$. Prior work, such as Meshry~\etal~\cite{Meshry2019NeuralRI}, represents such variation by learning an appearance vector from the exemplar image and a deep buffer containing semantic and depth information. However, their deep buffer is aligned with the viewpoint of the exemplar image. 
This makes the encoding of exemplar data view-dependent when, under fixed conditions, the information (e.g., sun direction) it reflects should 
be largely view-independent. 
In contrast, we utilize a deep buffer %
\emph{aligned with the reference viewpoint}. 

In particular, our encoder $E$ computes a latent appearance vector $\lv$:
\begin{align}
    \lv = E \left(\sourcei, \deepbuffer \right) 
\end{align}
where $\sourcei$ is an exemplar image and $\deepbuffer$ is a reference viewpoint--aligned deep buffer containing (1) a rectified RGB image over-composited from a PSV that reprojects exemplar $\sourcei$ to the reference viewpoint via the depth planes of the reference \deepmpi, (2) a flattened \albedo image over-composited from base color layers $\albedor$, and (3) a flattened latent feature map at the reference viewpoint over-composited from \deepmpi features $F^r$. 

\begin{figure}[t!]
    \centering
    \includegraphics[width=\columnwidth]{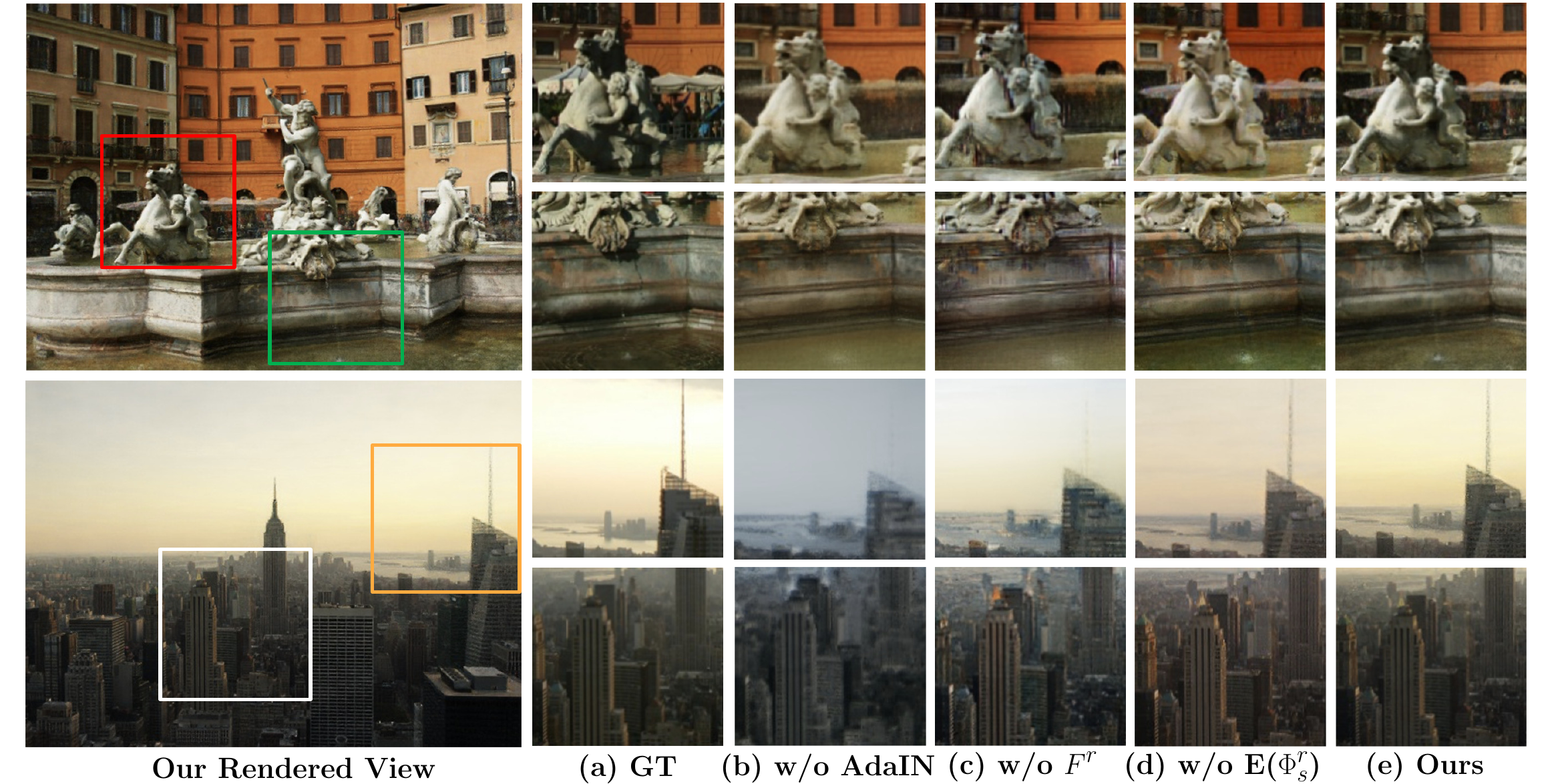} \\
    \caption{\textbf{Comparisons of images reconstructed with different configurations of our method.}  The images rendered from our full approach (e) are more similar to the ground truth images (a) than other configurations. In particular, the images rendered from the models without AdaIN (b) or the \deepmpi (c) are less realistic, and the model that does not feed the deep buffer $\deepbuffer$ to the encoder (d) fails to capture accurate scene appearance, as indicated in the highlighted regions. \label{fig:configurations}}
\end{figure}

Such a deep buffer allows $E$ to learn complex appearance by aligning the illumination information in the exemplar image with the shared scene intrinsic properties encoded in the reference \deepmpi. Without such alignment, it is difficult for $E$ to  consistently establish appearance correspondence across different viewpoints. Column (d) of Figure~\ref{fig:configurations} shows examples of rendered images without use of such a deep buffer. 
One can see that the deep buffer 
guides the model to capture complex illumination effects such as the realistic shadows highlighted in the first row. Moreover, integrating the \albedo and latent feature map at the reference viewpoint into $\deepbuffer$, and adding $\sourcei$ as inputs to $E$ can help the model to extrapolate appearance outside the field of view of the exemplar image, as shown in the last row of Figure~\ref{fig:app_transfer}.

\medskip
\noindent\textbf{Neural renderer.}
A plenoptic slice is now represented by the reference \deepmpi $\rfmpi$ and an appearance vector $\lv$.
Given these inputs, our neural renderer $G$ predicts the corresponding RGB color planes.
We could either predict these RGB planes at the reference viewpoint, or after first warping the DeepMPI to the target viewpoint. We choose the latter because it simplifies efficient implementation, as noted below.
Let $\kfmpi$ denote the reference DeepMPI $\rfmpi$ after warping into target viewpoint $c_k$. 
Given $\kfmpi$ and $\lv$, $G$ predicts the RGB color planes $\predmpi$ of a standard RGB$\alpha$ MPI at target viewpoint $c_k$:
\begin{align}
    \predmpi = G\left( \kfmpi,\ \lv \right)
\end{align}
In particular, $G$ takes in each layer of $\kfmpi$ \emph{independently} and predicts a corresponding RGB layer whose appearance is controlled by $\lv$. A rendered RGB image with the appearance of $I_s$ at viewpoint $c_k$ can then be obtained using the over operation with precomputed alpha weights $\alpha^k$ in $\kfmpi$:
\begin{align}
     \hat{I}^k_s = \mathcal{O}\left(\predmpi, \alpha^k\right)
 \end{align}
As shown in Fig.~\ref{fig:pipeline}, during training we set exemplar image $\sourcei = I_k$, i.e., we aim to reconstruct image $I_k$ at viewpoint $c_k$. At inference, $\sourcei$ is not necessarily $I_k$.

Our rendering network $G$ is a U-Net variant with an encoder-decoder architecture. Prior methods~\cite{Meshry2019NeuralRI,zhu2017toward} embed $\mathbf{z}$ in the bottleneck or input of $G$. Instead, we use Adaptive Instance Normalization (AdaIN) layers~\cite{huang2017arbitrary} whose parameters are dynamically generated from $\mathbf{z}$ via an MLP.
AdaIN 
has been shown to be effective in capturing both global and spatially varying appearance of exemplar images. 
We find that AdaIN not only helps model natural scene appearance, but also stabilizes training. Column (b) of Figure~\ref{fig:configurations} shows examples of our rendered images without AdaIN; one can see the model using AdaIN preserves more faithful scene appearance including the style and color of exemplar images.

In practice, feeding a full-resolution DeepMPI into $G$ and performing back-propagation is very memory intensive. Hence, during training, we operate on random $256\times 256$ crops of training images, and only the necessary portion of $\rfmpi$ is warped to $c_k$ and fed to $G$. At test time, any size input can be used.

\medskip
\noindent\textbf{Losses.}
To train $G$ and $E$, we compute losses between output views and ground-truth exemplar views. Our training loss is composed of three terms:
\begin{align}
    \mathcal{L} = \Lvgg + \wgan \Lgan + \wstyle \Lstyle,
\end{align}
where $\Lvgg$, $\Lgan$, and $\Lstyle$ denote VGG perceptual loss, adversarial loss, and style loss.
For $\Lvgg$, we adopt the 
formulation 
of \cite{zhou2018stereo,chen2017photographic}; $\Lgan$ is computed from multi-scale discriminators~\cite{wang2018high} with an objective similar to LSGAN~\cite{mao2017least}.

To further enforce that the appearance of rendered images matches that of exemplar images, our style loss $\Lstyle$ compares $l_1$ differences between Gram matrices constructed from VGG features at different layers. We empirically observe $\Lstyle$ can guide our model to correctly capture the appearance of exemplar images, especially for rare photos
such as those taken at sunset. 

%% file: 04-experiments.tex
\section{Experiments}

We conduct extensive experiments to validate our proposed approach on our Internet photo dataset. We first compare with two baseline methods both quantitatively and qualitatively on the tasks of view synthesis, appearance transfer and appearance interpolation. We also present an ablation study to examine the impact of different configurations of our model. Finally, we perform a user study whose results demonstrate the %
quality of our synthesized novel views.

\medskip
\noindent\textbf{Data and implementation.} We evaluate our approach on five of our reconstructed scenes, which contain on average 2,064 images. For each scene, images are randomly split into training and test sets with a 85:15 ratio. We train a separate model for each scene. %
To mask out transient objects such as people and cars during training and evaluation, we adopt state-of-the-art semantic and instance segmentation algorithms~\cite{chen2017rethinking,he2017mask} to create binary object masks. We set the dimension of the latent appearance vector to $\lv \in \mathbb{R}^{16}$, and that of our latent \deepmpi features to ${F}^r_\mathbf{p} \in \mathbb{R}^{8}$. We refer readers to the supplemental material for scene statistics, network architectures, and other implementation details.

\medskip
\noindent\textbf{Baselines.}
We compare our approach to two state-of-the-art multi-modal image-to-image translation methods, adapted to our task: MUNIT~\cite{huang2018multimodal} and Neural Rerendering in the Wild (NRW)~\cite{Meshry2019NeuralRI}. To compare to MUNIT, we adapt their network $G$ to predict an RGB image at the target viewpoint from a corresponding base color input, and train with a bidirectional reconstruction loss. For NRW, both $E$ and $G$ take as input \albedo, per-frame depth derived from the \deepmpi, and semantic segmentation at the target viewpoint. $G$ then predicts a corresponding RGB image conditioned on the appearance vector extracted by $E$. We follow the same staged training strategy and use the same losses as in~\cite{Meshry2019NeuralRI}.

\medskip
\noindent\textbf{Error metrics.}
Similar to \cite{Meshry2019NeuralRI}, we report test image reconstruction errors 
using three error metrics: $l_1$ error, peak signal-to-noise ratio (PSNR), and perceptual similarity 
(via LPIPS~\cite{zhang2018unreasonable}). Prior work has found the LPIPS metric 
to be better correlated with human visual perception than other metrics.

\input{tables/tab-quant}

\begin{figure}[t!]
  \centering
    \begin{tabular}{@{}c@{}c@{}c@{}c@{}}
        \includegraphics[width=0.24\columnwidth]{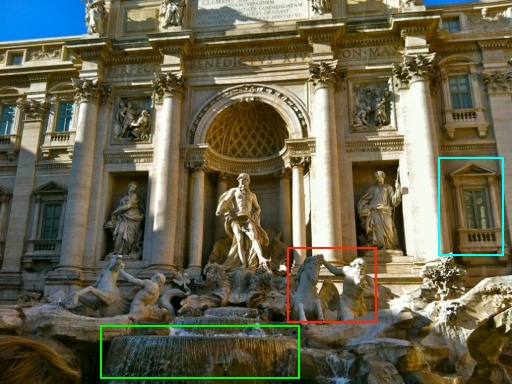}  &
        \includegraphics[width=0.24\columnwidth]{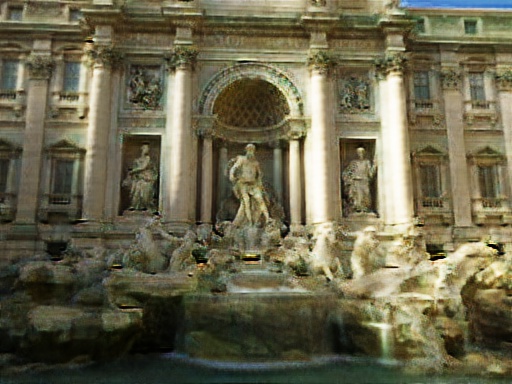}  &
        \includegraphics[width=0.24\columnwidth]{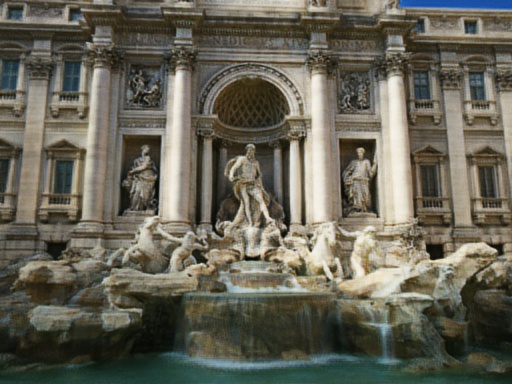}  & 
        \includegraphics[width=0.24\columnwidth]{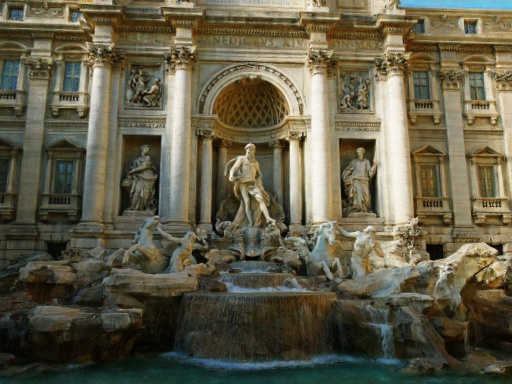}
         \\
        \includegraphics[width=0.24\columnwidth]{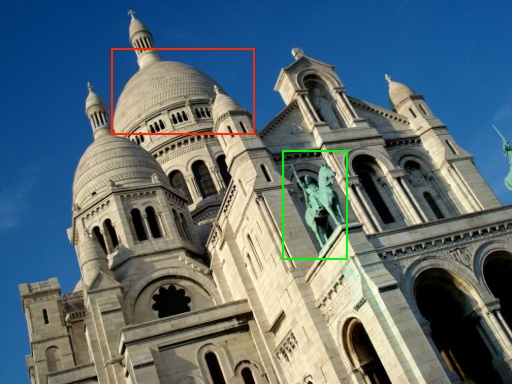}      &
        \includegraphics[width=0.24\columnwidth]{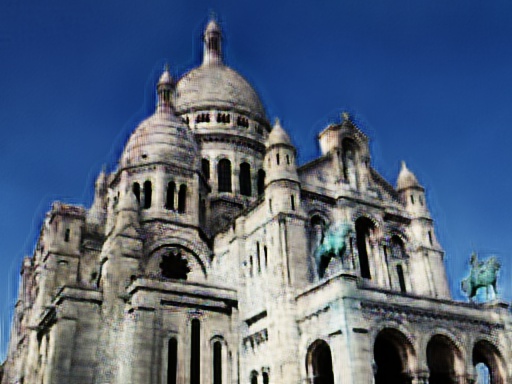}  &
        \includegraphics[width=0.24\columnwidth]{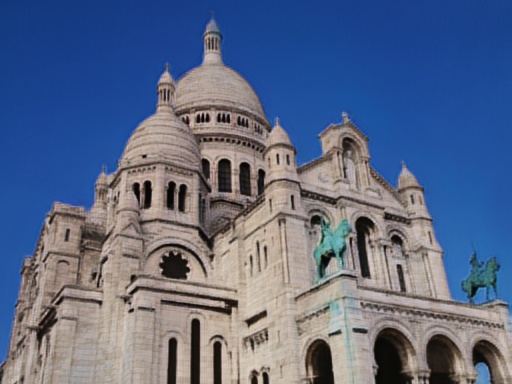}  & 
        \includegraphics[width=0.24\columnwidth]{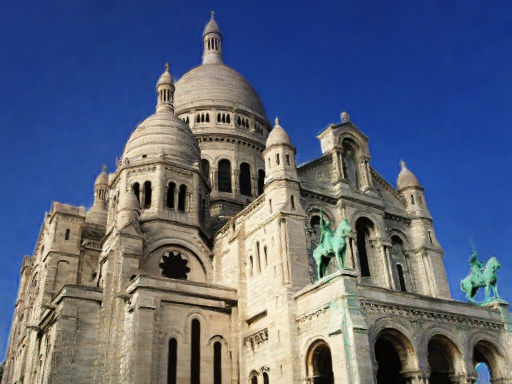} \\
        \includegraphics[width=0.24\columnwidth]{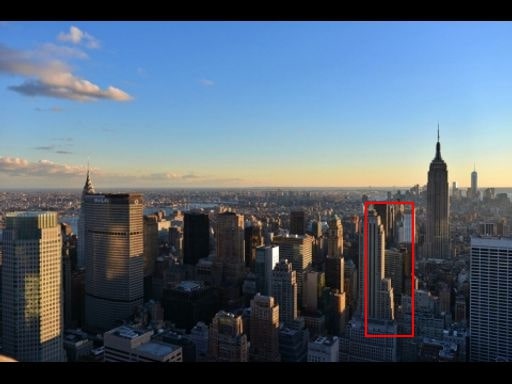}  &
        \includegraphics[width=0.24\columnwidth]{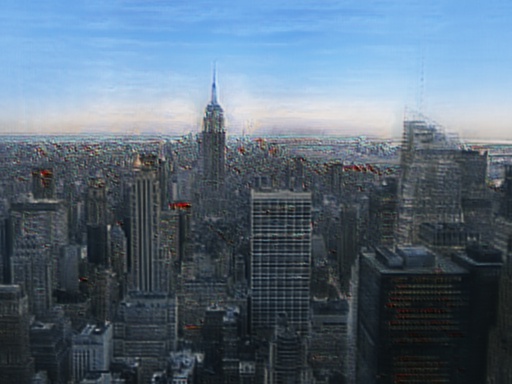}  &
        \includegraphics[width=0.24\columnwidth]{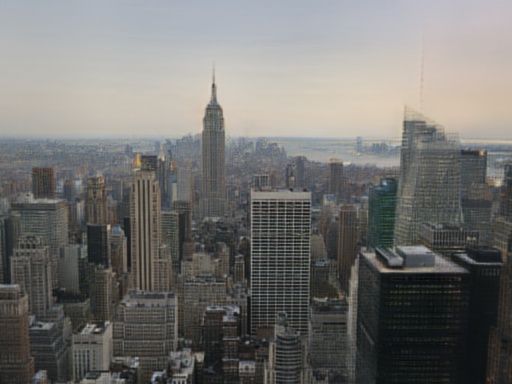}  & 
        \includegraphics[width=0.24\columnwidth]{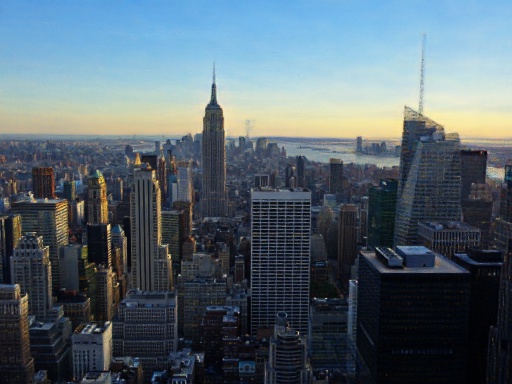} \\
        \includegraphics[width=0.24\columnwidth]{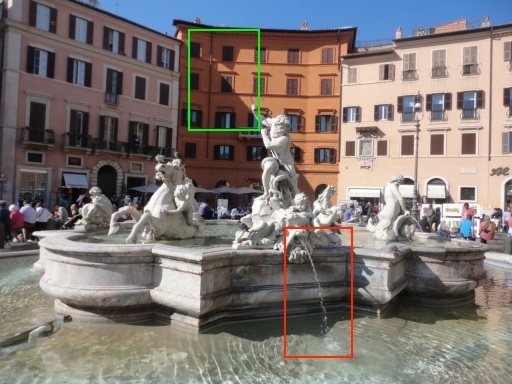}  &
        \includegraphics[width=0.24\columnwidth]{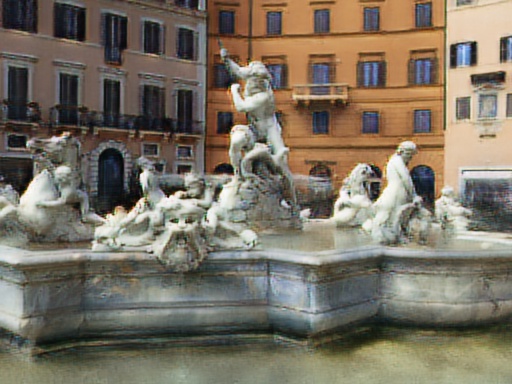}  &
        \includegraphics[width=0.24\columnwidth]{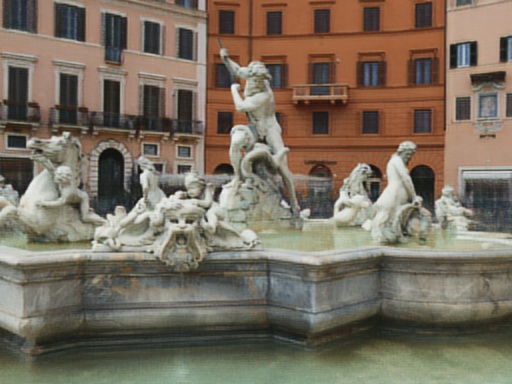}  & 
        \includegraphics[width=0.24\columnwidth]{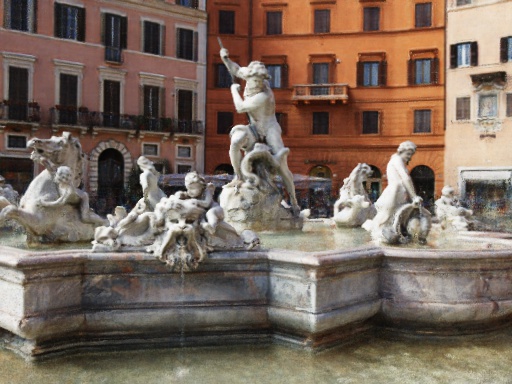} \\
        \includegraphics[width=0.24\columnwidth]{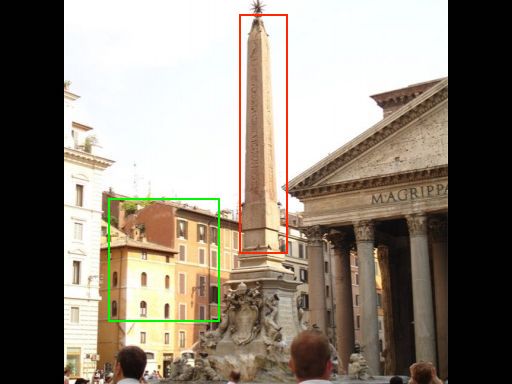}  &
        \includegraphics[width=0.24\columnwidth]{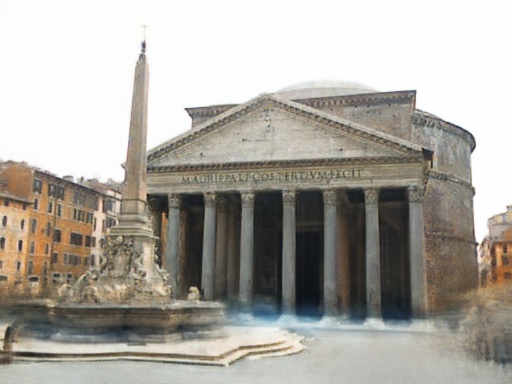}  &
        \includegraphics[width=0.24\columnwidth]{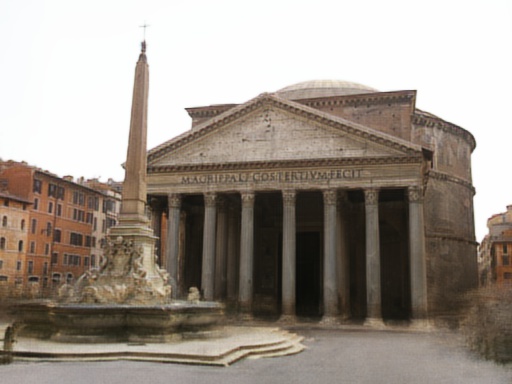}  & 
        \includegraphics[width=0.24\columnwidth]{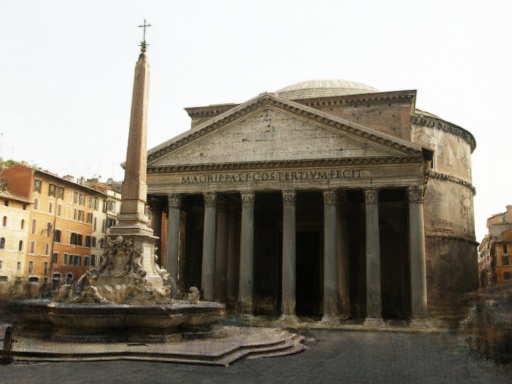} \\
        (a) exemplar $I$  & 
        (b) MUNIT~\cite{huang2018multimodal}  &
        (c) NRW~\cite{Meshry2019NeuralRI}  &
        (d) Ours  \\
    \end{tabular}
  	\caption{\textbf{Appearance transfer comparison.} From left to right: (a) exemplar images used to extract appearance vectors, (b) predictions from  MUNIT~\cite{huang2018multimodal}, (c) predictions from NRW~\cite{Meshry2019NeuralRI}, (d) predictions from our method. Compared to the baselines, our rendered images are more photo-realistic and are more faithful to the appearance of the exemplar images. Please zoom in to highlighted regions for better visual comparisons. } \label{fig:app_transfer}
 \end{figure}

\medskip
\noindent\textbf{Quantitative comparison.}
For fair comparison, we train and evaluate the baselines using the same data and hyperparameter settings as our method. Table~\ref{tb:quantitative} shows results of quantitative comparisons on our test set. Our proposed approach outperforms the two baseline methods by a large margin in terms of $l_1$ and PSNR, and is significantly better in terms of LPIPS, indicating that our method achieves higher rendering quality and realism. 

\medskip
\noindent\textbf{Ablation study.}
We perform an ablation study to 
analyze the effect
of individual system components. In particular, we replace four
components 
with simpler configurations: (1) using a train-from-scratch baseline to estimate alpha, as described in Section~\ref{sec:mpi} (w/o 2-phase), (2) including $\mathbf{z}$ as an input to $G$ rather than using AdaIN (w/o AdaIN), (3) removing latent features from the \deepmpi (w/o $F^r$), and (4) encoding $\mathbf{z}$ only from the exemplar image and not additionally from the deep buffer (w/o $E(\deepbuffer)$). Quantitative results are reported in Table~\ref{tb:quantitative}. 
Latent \deepmpi features, as well as the use of AdaIN in our neural renderer, yield significant improvements, and lead to better rendering quality for thin structures and attached shadows, as highlighted in Figure~\ref{fig:configurations}. Encoding the reference deep buffer also yields rendered images that better match the exemplar image.

\medskip
\noindent\textbf{Rendering with appearance transfer.}
Figure~\ref{fig:app_transfer} shows qualitative comparisons between our method and the two baselines on our test set in terms of rendering quality and appearance transferability (i.e., how well the model can transfer illumination and appearance of an exemplar image to a 
target viewpoint). We demonstrate compelling results in challenging cases such as sunset, which is a rare condition in the input photos. Compared to MUNIT, our rendered images are more realistic and exhibit fewer artifacts. For example, our rendered images successfully model specularities on glass windows, details on running water and droplets, cast shadows, and directional lighting effect as shown in the highlighted regions in Figure~\ref{fig:app_transfer}. Our approach can also generate complex highlights and cast shadows from the sun.
Compared with NRW, our rendered images are more faithful to the illumination in the exemplar image (e.g., for sunset appearance). Moreover, our approach can extrapolate appearance beyond the field of view of the exemplar image, as shown in the last row of Figure~\ref{fig:app_transfer}. We refer readers to the supplemental video for visual comparisons with animated camera trajectories.

\begin{figure}[t!]
    \centering
    \includegraphics[width=\columnwidth]{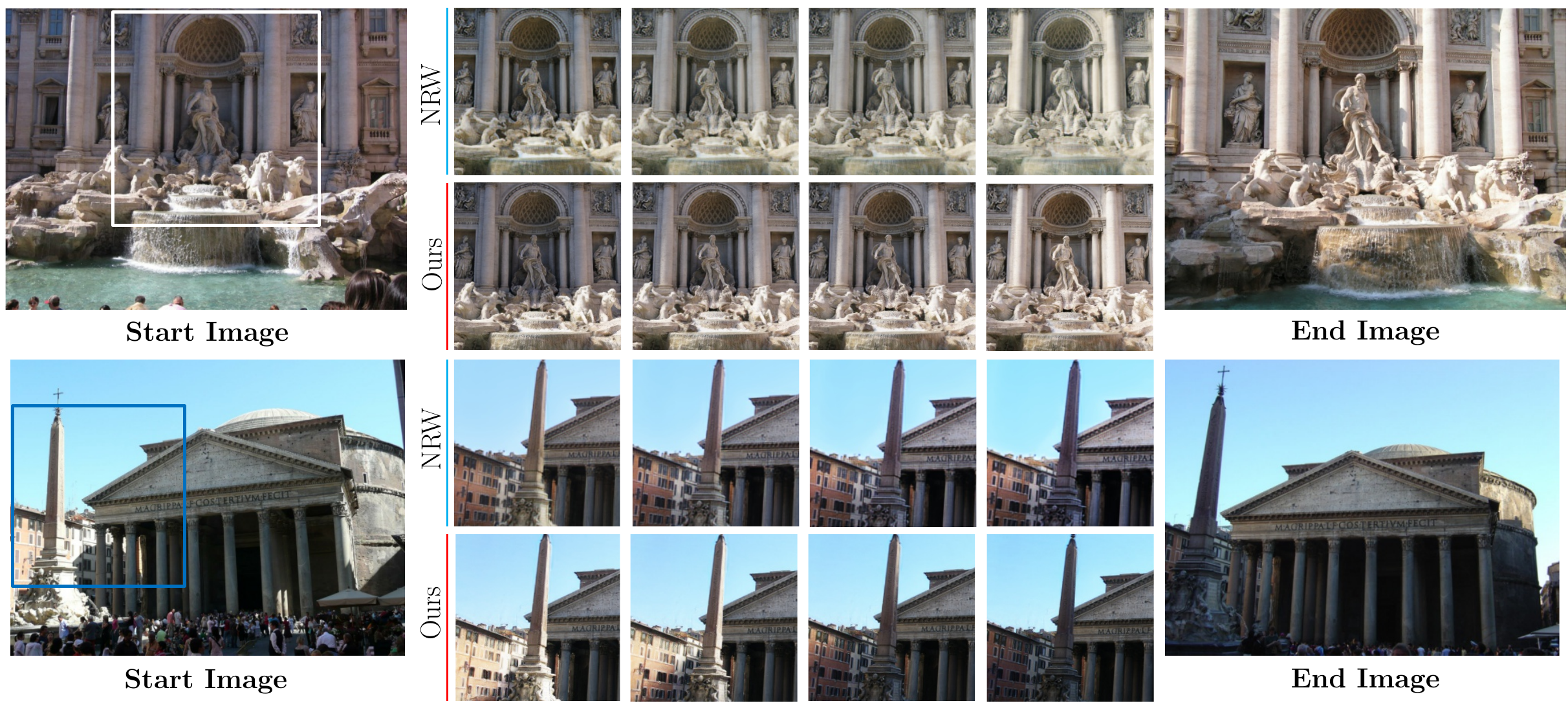} \\
    \caption{\textbf{Appearance interpolation.} The left- and rightmost exemplar images indicate start and end appearance. Intermediate images are generated by linearly interpolating latent vectors from the two images. Odd rows show interpolation results from NRW~\cite{Meshry2019NeuralRI}, and even rows from our method. Moving shadows are indicated in highlighted regions.} \label{fig:app_interpolation}
\end{figure}

\medskip
\noindent\textbf{Appearance interpolation.}
A key advantage of our method is the ability to interpolate between plenoptic slices in the latent appearance space. 
We conduct qualitative comparisons between our approach and NRW on appearance interpolation. %
We choose two images 
to define the start and end appearance, and linearly interpolate their latent vectors to produce in-between appearances. Figure~\ref{fig:app_interpolation} shows a comparison of interpolation results. In the first two rows of the figure,
we observe that our method can simulate the progression of surfaces exposed to sunlight as the sun moves, while NRW fails to produce this effect. In the last row, our approach recovers the gradual motion of shadows throughout the day, while shadows in the NRW results tend to fade less naturally during interpolation. We refer readers to the supplemental videos for animated comparisons.

\begin{figure}[t!]
  \centering
    \begin{tabular}{@{}c@{}c@{}c@{}c@{}c@{}c@{}c@{}}
        \includegraphics[width=0.14\columnwidth]{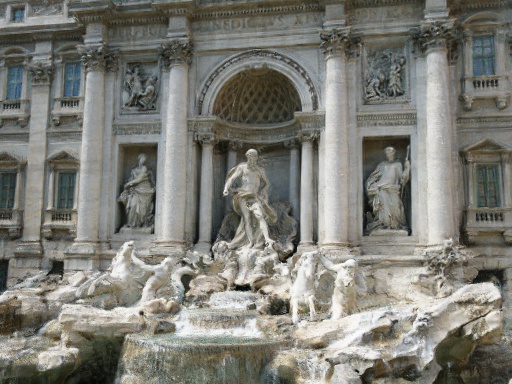}  & 
        \includegraphics[width=0.14\columnwidth]{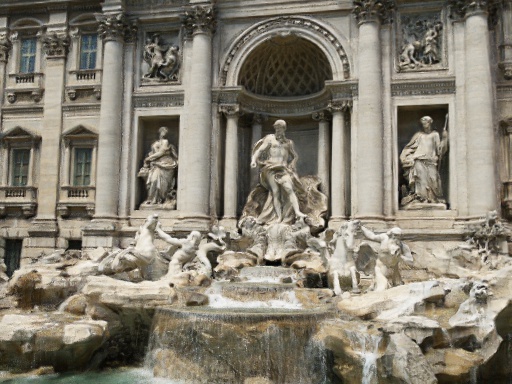}  & 
        \includegraphics[width=0.14\columnwidth]{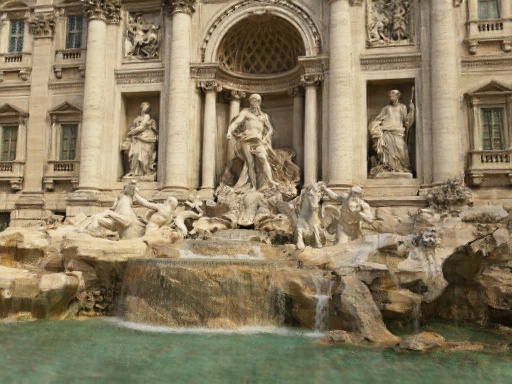}  & 
        \includegraphics[width=0.14\columnwidth]{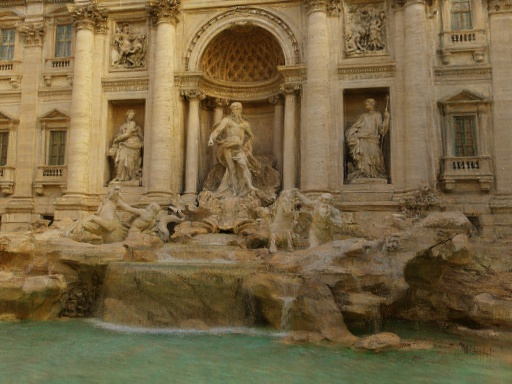}  &
        \includegraphics[width=0.14\columnwidth]{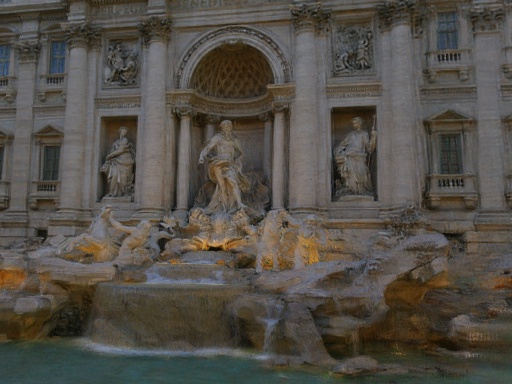}  & 
        \includegraphics[width=0.14\columnwidth]{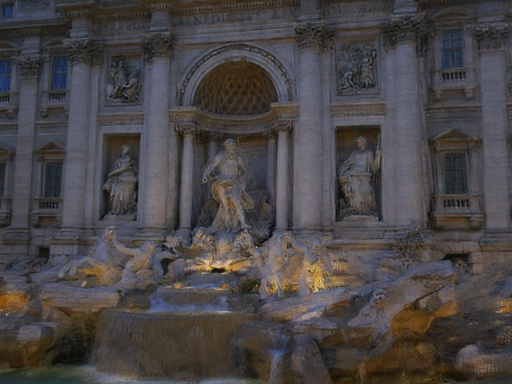}  &
        \includegraphics[width=0.14\columnwidth]{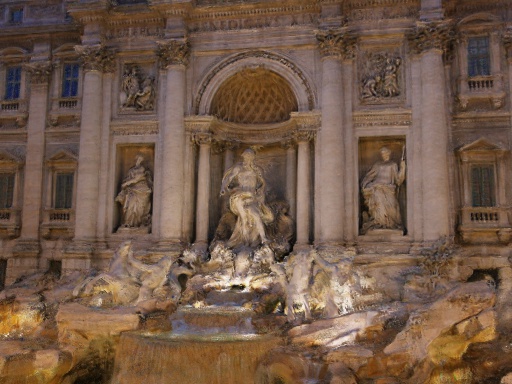}  \\
        \includegraphics[width=0.14\columnwidth]{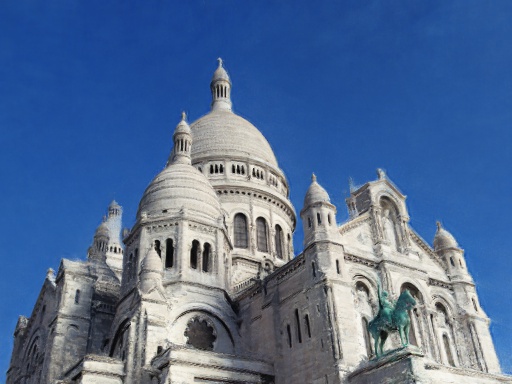}  & 
        \includegraphics[width=0.14\columnwidth]{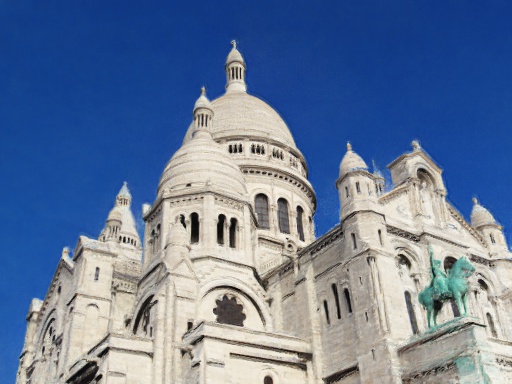}  & 
        \includegraphics[width=0.14\columnwidth]{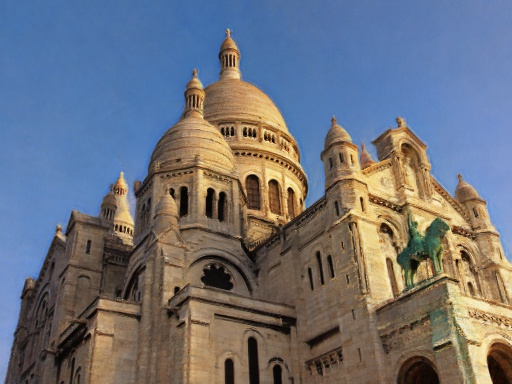}  & 
        \includegraphics[width=0.14\columnwidth]{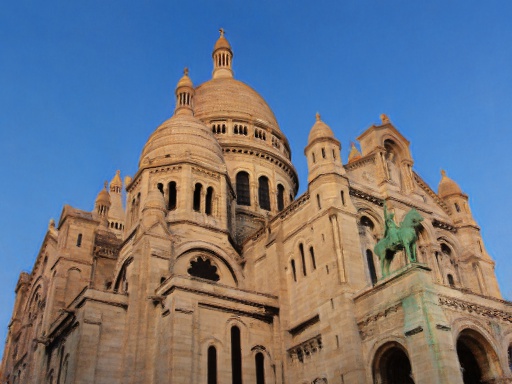}  &
        \includegraphics[width=0.14\columnwidth]{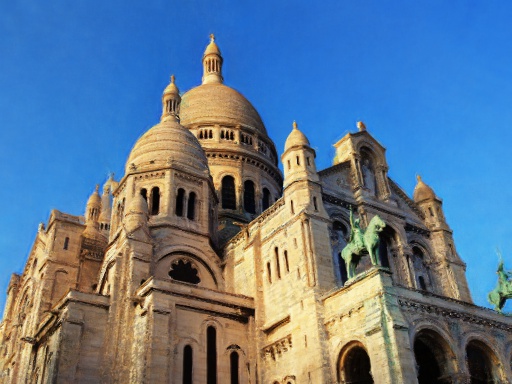}  & 
        \includegraphics[width=0.14\columnwidth]{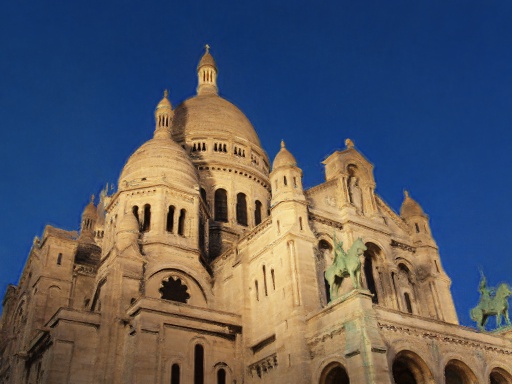}  &
        \includegraphics[width=0.14\columnwidth]{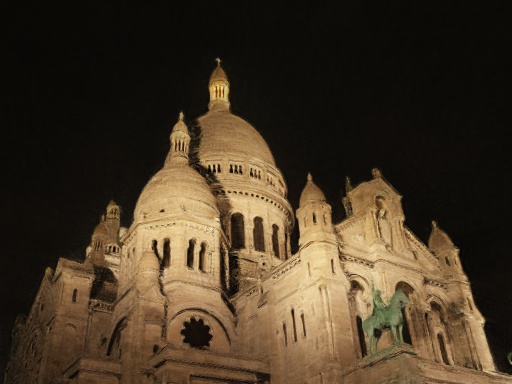}  \\
    \end{tabular}
    \caption{\textbf{4D Photos}. We demonstrate an application of creating \emph{4D photos} by performing spatial-temporal interpolation in which both camera viewpoint and scene illumination change simultaneously. Results are best appreciated in the supplementary videos.} \label{fig:app_4d}
 \end{figure}

\medskip
\noindent\textbf{4D photos.}
Figure~\ref{fig:app_4d} shows an application of our method to generating animated \emph{4D photos} by animating the 3D viewpoints and simultaneously interpolating between latent appearance features. Our results achieve convincing changes across a variety of times of day and lighting conditions. The parallax effect of our results is best appreciated in the supplemental videos. 

\medskip
\noindent\textbf{User study.} 
We ran a user study using 24 random sets of videos with camera movements and synthesized images from 5 different scenes. Each video is a sequence of novel views generated by our method, NRW~\cite{Meshry2019NeuralRI}, or MUNIT~\cite{huang2018multimodal}. To quantify the performance of appearance transfer, we also show comparisons of results generated from different exemplar images selected from our test set. We invited 46 participants and asked them to rank the results of the three approaches. 88\% of the time, participants responded that the videos produced by our system are the most temporally coherent. 82\% of the time, they responded that the results from our method best reproduce the details of structure and illumination one would expect of a real-world scene. 77\% of the time, they responded that the results from our method are the most faithful to 
the corresponding exemplar.

%% file: tables/tab-quant.tex
\begin{table*}[t!]
\centering
{
\resizebox{\textwidth}{!}{
\begin{tabular}{lcccccccccccccccc}
\toprule
 &  \multicolumn{3}{c}{\textbf{Trevi Fountain}} & \multicolumn{3}{c}{\textbf{Sacre Coeur}} & \multicolumn{3}{c}{\textbf{The Pantheon}} & \multicolumn{3}{c}{\textbf{Top of the Rock}} & \multicolumn{3}{c}{\textbf{Piazza Navona}} \\
 
     \textbf{Method}  & $l_1$ & \text{LPIPS}  & \text{PSNR} & $l_1$ & \text{LPIPS} & \text{PSNR} & $l_1$ & \text{LPIPS} & \text{PSNR} & $l_1$ & \text{LPIPS} & \text{PSNR} & $l_1$ & \text{LPIPS} & \text{PSNR} \\
\midrule
MUNIT~\cite{huang2018multimodal}  & 0.768 & 2.62 & 20.1 & 0.740 & 2.08 & 20.2 & 0.560 & 1.51 & 21.4 & 0.876 & 3.68 & 18.2 & 0.984 & 2.80 & 17.4 \\
NRW~\cite{Meshry2019NeuralRI}  & 0.779 & 2.07 & 20.0 & 0.808 & 1.90 & 19.6 & 0.592 & 1.35 & 21.1 & 0.802 & 2.76 & 19.3 & 1.050 & 2.64 & 17.1 \\
\midrule 
w/o 2-phase & 0.651	& 1.68	& 21.0 & 0.695 & 1.61 & 20.8 & 0.515 & 1.12 & 21.9 & 0.694 & 2.19 & 20.4 &	1.010  & 2.52 & 17.4 \\
w/o AdaIN  & 0.780 & 1.87 & 19.8 & 0.801 & 1.89	& 19.6 & 0.609 & 1.30 & 20.9 & 0.773 & 2.58 & 19.3 &	1.150 & 2.97  & 17.1 \\
w/o $F^r$ & 0.712 & 1.74 & 20.5 & 0.737 & 1.78 & 20.2 & 0.556 & 1.25 & 21.5 & 0.720 & 2.47 & 19.9 & 1.045 & 2.62 & 17.0 \\
w/o $E(\deepbuffer)$  & 0.670 & 1.70 & 20.9 &	 0.715 & 1.66 & 20.5 & 0.549 & 1.16 & 21.5 & 0.703 & 2.24 & 20.0 & 1.017 & 2.52 & 17.2 \\
Ours (full)  & \textbf{0.618} & \textbf{1.56} & \textbf{21.8} & \textbf{0.676} & \textbf{1.57} & \textbf{21.0} & \textbf{0.495} & \textbf{1.08} & \textbf{22.5} & \textbf{0.642} & \textbf{2.48} & \textbf{20.7} & \textbf{0.933} & \textbf{2.32} & \textbf{17.6} \\
\bottomrule
\end{tabular}}
}
\caption{{\bf Quantitative comparisons on our test set.} Lower is better for $l_1$ and LPIPS and higher is better for PSNR. $l_1$ errors are scaled by 10 for ease of presentation.} \label{tb:quantitative}
\end{table*}

%% file: 06-conclusion.tex
\section{Discussion and Conclusion}

\begin{figure}[t!]
  \centering
    \begin{tabular}{@{}c@{}c@{}c@{}c@{}}
        \includegraphics[width=0.245\columnwidth]{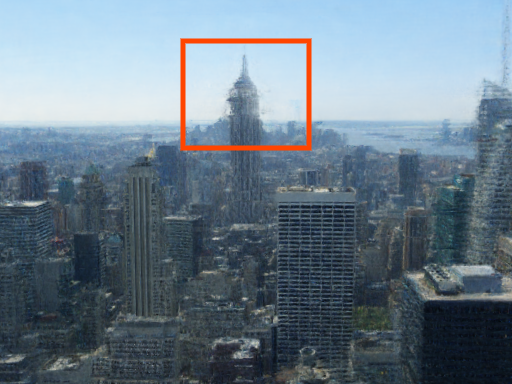} & 
        \includegraphics[width=0.245\columnwidth]{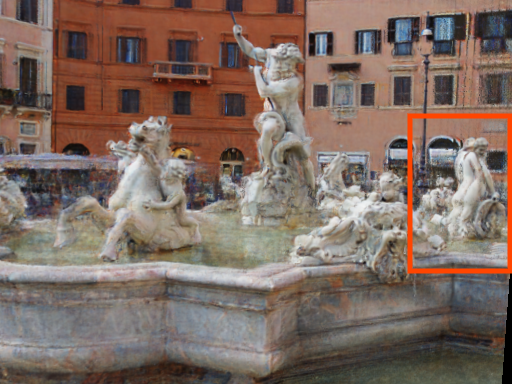} & 
        \includegraphics[width=0.245\columnwidth]{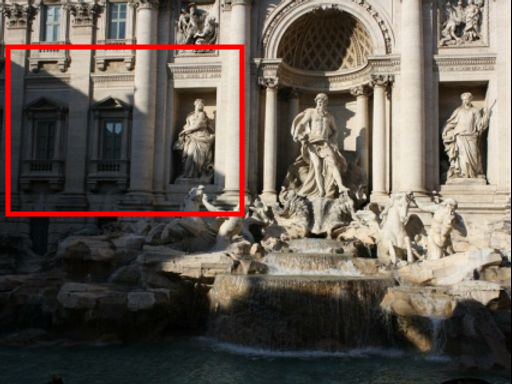} &
        \includegraphics[width=0.245\columnwidth]{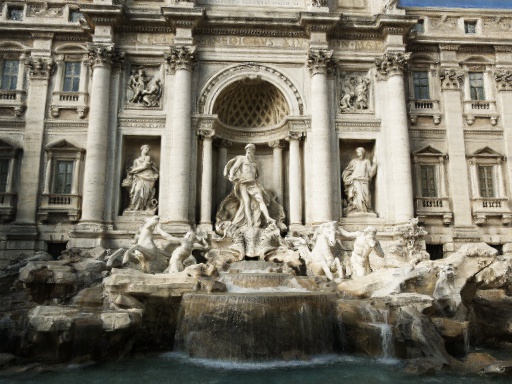} \\
        (a) insufficient view  & 
        (b) MPI limits  &
        (c) exemplar for (d)  &
        (d) missing shadow  \\
    \end{tabular}
    \caption{\textbf{Limitations.} Some failure cases include: (a) input photo collections that do not span the full range of desired viewpoints, or (b) intrinsic limitations of MPI leading to poor extrapolation to large camera motions. In addition, as shown in (c) (exemplar image with strong shadow) and (d) (resulting rendering), our method can fail to model strong cast shadows produced by occluders outside the reference field of view.   \label{fig:limitations}}
 \end{figure}

\noindent \textbf{Limitations.} Our method inherits limitations from 
MPIs.
For example, MPIs fail to generalize to viewpoints that are not well-sampled, or that are far from the reference view of the MPI (see Figure~\ref{fig:limitations}(a-b)).
In addition, our model can also sometimes fail to model cast shadows from occluders outside of the reference field of view, as shown in Figure~\ref{fig:limitations}(c) and (d). Despite these limitations, we believe our work represents a significant advance towards photo-realistic capture and rendering of the world from crowd photography.

\medskip
\noindent \textbf{Conclusion.} We presented a method for synthesizing novel views of scenes under time-varying appearance from Internet photos. We proposed a new \deepmpi representation and a method for optimizing and decoding DeepMPIs conditioned on viewing conditions present in different photos. 
Our method can synthesize
plenoptic slices that can be interpolated to recover local regions of the full plenoptic function. In the future, we 
envision
enabling even larger changes in viewpoint and illumination, including 4D walkthroughs of large-scale scenes.

\medskip
{\small
\noindent \textbf{Acknowledgements.} 
We thank Kai Zhang, Jin Sun, and Qianqian Wang for helpful discussions.
This research was supported in part by the generosity of Eric and Wendy Schmidt by recommendation of the Schmidt Futures program.
}